%% file: main.tex
\documentclass[11pt,a4paper]{article}
\usepackage{acl2014}
\usepackage{times}
\usepackage{url}
\usepackage{latexsym}

\usepackage{booktabs}
\usepackage{amsmath}
\usepackage{multirow}

\newcommand{\ov}{\overrightarrow}

\title{Evaluating Neural Word Representations in\\Tensor-Based Compositional Settings}

\author{Dmitrijs Milajevs$^1$~~~~Dimitri Kartsaklis$^2$~~~~Mehrnoosh Sadrzadeh$^1$~~~~Matthew Purver$^1$\vspace{0.2cm}\\
\begin{minipage}{0.50\linewidth}
\centering
$^1$Queen Mary University of London\\School of Electronic Engineering\\and Computer Science\\
Mile End Road, London, UK
\texttt{\small \{d.milajevs,m.sadrzadeh,m.purver\}@qmul.ac.uk}
\vspace{0.2cm}
\end{minipage}
\begin{minipage}{0.50\linewidth}
\centering
$^2$University of Oxford\\
Department of Computer Science\\
Parks Road, Oxford, UK
\texttt{\small dimitri.kartsaklis@cs.ox.ac.uk}\\
~
\end{minipage}
}

\begin{document}
\maketitle

\begin{abstract}
  We provide a comparative study between neural word
  representations and traditional vector spaces based on co-occurrence
  counts, in a number of compositional tasks. We use three different
  semantic spaces and implement seven tensor-based compositional
  models, which we then test (together with simpler additive and
  multiplicative approaches) in tasks involving verb disambiguation
  and sentence similarity. To check their scalability, we additionally
  evaluate the spaces using simple compositional methods on
  larger-scale tasks with less constrained language: paraphrase
  detection and dialogue act tagging. In the more constrained tasks,
  co-occurrence vectors are competitive, although choice of compositional method is
  important; on the larger-scale
  tasks, they are outperformed by neural word embeddings, which
  show robust, stable performance across the tasks.
\end{abstract}

\section{Introduction}
\label{sec:introduction}

Neural word embeddings \cite{bengio2006,collobert2008,mikolov2013efficient} have
received much attention in the distributional semantics community, and have
shown state-of-the-art performance in many natural language processing
tasks. While they have been compared with co-occurrence based models in simple
similarity tasks at the word level \cite{levy2014linguistic,baroni2014don}, we
are aware of only one work that attempts a comparison of the two approaches in
compositional settings \cite{blacoe2012comparison}, and this is limited
to additive and multiplicative composition, compared against composition via a
neural autoencoder.

The purpose of this paper is to provide a more complete picture
regarding the potential of neural word embeddings in compositional
tasks, and meaningfully compare them with the traditional
distributional approach based on co-occurrence counts. We are
especially interested in investigating the performance of neural word
vectors in compositional models involving general mathematical
composition operators, rather than in the more task- or
domain-specific deep-learning compositional settings they have
generally been used with so far (for example, by \newcite{socher2012semantic},
\newcite{kalchbrenner-blunsom2013CVSC} and many others).

In particular, this is the first large-scale study to date that
applies neural word representations in tensor-based compositional
distributional models of meaning similar to those formalized by
\newcite{coecke2010}. We test a range of implementations based on this
framework, together with additive and multiplicative approaches
\cite{mitchell2008vector}, in a variety of different
tasks. Specifically, we use the verb disambiguation task of
\newcite{grefenstette2011experimental} and the transitive sentence
similarity task of \newcite{KartSadrQPL} as small-scale focused
experiments on pre-defined sentence structures. Additionally, we
evaluate our vector spaces on paraphrase detection (using the
Microsoft Research Paraphrase Corpus of \newcite{dolan2005microsoft})
and dialogue act tagging using the Switchboard Corpus
(see e.g.~\cite{Stolcke.etal00}).

In all of the above tasks, we compare the neural word embeddings of
\newcite{mikolov2013efficient} with two vector spaces both based on
co-occurrence counts and produced by standard distributional
techniques, as described in detail below. The general picture we get
from the results is that in almost all cases the neural vectors are
more effective than the traditional approaches.

We proceed as follows: Section \ref{sec:meaning-representation} provides a
concise introduction to distributional word representations in natural language
processing. Section \ref{sec:compositional-models} takes a closer look to the
subject of compositionality in vector space models of meaning and describes the
range of compositional operators examined here. In Section
\ref{sec:semantic-spaces} we provide details about the vector spaces used in the
experiments. Our experimental work is described in detail in Section
\ref{sec:experiments}, and the results are discussed in Section
\ref{sec:discussion}. Finally, Section \ref{sec:concl-future-work} provides conclusions.

\section{Meaning representation}
\label{sec:meaning-representation}

There are several approaches to the representation of word, phrase and
sentence meaning. As natural languages are highly creative and it is
very rare to see the same sentence twice, any practical approach dealing with large text segments must
be \emph{compositional}, constructing the meaning of phrases and
sentences from their constituent parts. The ideal method would
therefore express not only the similarity in meaning between those
constituent parts, but also between the results of their composition, and do this in
ways which fit with linguistic structure and generalisations thereof.

\paragraph{Formal semantics}
\label{sec:formal-semantics}

Formal approaches to the semantics of natural language have long built
upon the classical idea of compositionality -- that the meaning of a
sentence is a function of the meanings of its parts \cite{frege1892sense}. In
compositional type-logical approaches, predicate-argument structures
representing phrases and sentences are built from their constituent
parts by $\beta$-reduction within the lambda
calculus framework \cite{montague1970universal}: for example, given a
representation of \emph{John} as $\mathit{john}'$ and
\emph{sleeps} as $\lambda x.\mathit{sleep}'(x)$, the meaning of the sentence
``John sleeps'' can be constructed as $\lambda
x.\mathit{sleep}'(x)(\mathit{john}') =
\mathit{sleep}'(\mathit{john}')$. Given a suitable pairing between words and semantic representations of them, this method can produce structured sentential
representations with broad coverage and good generalisability (see
e.g.~\cite{Bos2008STEP2}). The above logical approach is extremely powerful
because it can capture complex aspects of meaning such as quantifiers
and their interaction (see e.g.~\cite{copestake2005minimal}), and
enables inference using well studied and developed logical methods
(see e.g.~\cite{bos2000first}).

\paragraph{Distributional hypothesis}
\label{sec:distr-hypoth}

However, such formal approaches are less able to express
\emph{similarity} in meaning. We would like to capture the intuition
that while \textit{John} and \textit{Mary} are distinct, they are
rather similar to each other (both of them are humans) and dissimilar
to words such as \textit{dog}, \textit{pavement} or \textit{idea}. The same applies
at the phrase and sentence level: ``dogs chase cats'' is similar
in meaning to ``hounds pursue kittens'', but less so to
``cats chase dogs'' (despite the lexical overlap).

Distributional methods provide a way to address this problem. By
representing words and phrases as vectors or tensors in a (usually
highly dimensional) vector space, one can express similarity in meaning
via a suitable distance metric within that space (usually cosine distance); furthermore, composition can be modelled via suitable linear-algebraic operations.

\paragraph{Co-occurrence-based word representations}
\label{sec:distr-repr}

One way to produce such vectorial representations is to directly exploit
\newcite{harris1954distributional}'s intuition that semantically
similar words tend to appear in similar contexts. We can construct a
vector space in which the dimensions correspond to contexts, usually taken to be
 words as well. The word vector components can then be calculated from
the frequency with which a word has co-occurred with the corresponding
contexts in a  window of words, with a predefined length.

\begin{table}[b!]
  \centering
  \begin{tabular}{lrrr}
    \toprule
    & philosophy & book & school \\
    \midrule
    Mary & 0  & 10 & 22  \\
    John & 4  & 60 & 59  \\
    girl & 0  & 19 & 93  \\
    boy  & 0  & 12 & 164 \\
    idea & 10 & 47 & 39  \\
    \bottomrule
  \end{tabular}
  \caption{Word co-occurrence frequencies extracted from the BNC \cite{leech1994claws4}.}
  \label{tab:comparison}
\end{table}

Table~\ref{tab:comparison} shows 5 3-dimensional vectors for the words
\textit{Mary}, \textit{John}, \textit{girl}, \textit{boy} and \textit{idea}. The
words \textit{philosophy}, \textit{book} and \textit{school} signify vector space
dimensions. As the vector for \textit{John} is closer to \textit{Mary} than it is
to \textit{idea} in the vector space---a direct consequence of the fact that \textit{John}'s
contexts are similar to \textit{Mary}'s and dissimilar to
\textit{idea}'s---we can infer that \textit{John} is semantically more similar
to \textit{Mary} than to \textit{idea}.

Many variants of this approach exist: performance on word similarity
tasks has been shown to be improved by replacing raw counts with
weighted values (e.g.~mutual information)---see
\cite{turney2010frequency} and below for discussion, and
\cite{kiela-clark:2014:CVSC} for a detailed comparison.

\paragraph{Neural word embeddings}
\label{sec:neural-embedding}

Deep learning techniques exploit the distributional hypothesis
differently. Instead of relying on observed co-occurrence frequencies,
a neural language model is trained to maximise some objective function related
to e.g. the probability of observing the surrounding words in some
context \cite{mikolov2013distributed}:
\begin{align}
 \frac{1}{T}\sum^{T}_{t=1}\sum_{-c \leq j \leq c, j\neq0} \log p(w_{t+j}|w_t)
  \label{eq:objective-func}
\end{align}
\noindent
Optimizing the above function, for example, produces vectors which maximise the
conditional probability of observing words in a context around the
target word $w_t$, where $c$ is the size of the training window, and
$w_1 w_2, \cdots w_T$ a sequence of words forming a training instance. Therefore, the resulting vectors
will capture the distributional intuition and can express degrees of
lexical similarity.

This method has an obvious advantage compared to co-occurrence method: since now the context is \textit{predicted}, the model in principle can be much more robust in data sparsity problems, which is always an important issue for co-occurrence word spaces.
Additionally, neural vectors have also proven successful in other tasks
\cite{mikolov2013linguistic}, since they seem to encode not only
attributional similarity (the degree to which similar words are close to each other), but
also relational similarity \cite{turney2006similarity}. For example,
it is possible to extract the singular:plural relation
(\textit{apple}:\textit{apples}, \textit{car}:\textit{cars}) using
vector subtraction:
\begin{align*}
  \overrightarrow{\mathit{apple}} - \overrightarrow{\mathit{apples}}
  \approx
  \overrightarrow{\mathit{car}} - \overrightarrow{\mathit{cars}}
\end{align*}
Perhaps even more importantly, semantic relationships are preserved in a very intuitive way:
\begin{align*}
  \overrightarrow{\mathit{king}} - \overrightarrow{\mathit{man}}
  \approx
  \overrightarrow{\mathit{queen}} - \overrightarrow{\mathit{woman}}
\end{align*}
allowing the formation of analogy queries similar to
$\overrightarrow{\mathit{king}} - \overrightarrow{\mathit{man}} +
\overrightarrow{\mathit{woman}} = \mathtt{?}$, obtaining
$\overrightarrow{\mathit{queen}}$ as the
result.\footnote{\newcite{levy2014linguistic} improved
  \newcite{mikolov2013linguistic}'s method of retrieving relational similarities
  by changing the underlying objective function.}

Both neural and co-occurrence-based approaches have advantages over
classical formal approaches in their ability to capture lexical
semantics and degrees of similarity; their success at extending this
to the sentence level and to more complex semantic phenomena, though, depends
on their applicability within compositional models, which is the subject of the next section.

\section{Compositional models}
\label{sec:compositional-models}

Compositional distributional models represent meaning of a sequence
of words by a vector, obtained by combining  meaning vectors of the
words within the sequence using some vector composition operation. In a
general classification of these models, one can distinguish between
three broad cases: simplistic models which combine word
vectors irrespective of their order or relation to one another, models
which exploit linear word order, and models which use grammatical
structure.

The first approach combines word vectors by vector addition or point-wise
multiplication \cite{mitchell2008vector}---as this is independent of
word order, it cannot capture the difference between the two sentences ``dogs
  chase cats'' and ``cats chase dogs''. The second approach
has generally been implemented using some form of deep learning, and
captures word order, but not by necessarily caring about the
grammatical structure of the sentence. Here, one works by recursively
building and combining vectors for subsequences of words within the
sentence using e.g.~autoencoders \cite{socher2012semantic} or
convolutional filters \cite{KalchbrennerACL2014}. We do not consider
this approach in this paper. This is because, as mentioned in the introduction,  their vectors and composition
operators are task-specific. These are trained directly to achieve
specific objectives in certain pre-determined tasks. We are
interested in vector and composition operators that work for \textit{any}
compositional task, and which can be combined with results
in linguistics and formal semantics to provide generalisable models that
can canonically extend to complex semantic phenomena.
The third (i.e. the grammatical) approach promises a way to achieve this, and
has been instantiated in various ways in the work of
\newcite{baroni2010nouns},\newcite{grefenstette2011experimental}, and \newcite{KartSadrCOLING}.

\paragraph{General framework}

Formally, we can specify the vector representation of a word sequence $w_1 w_2 \cdots w_n$ as the vector $\overrightarrow{s} =
\overrightarrow{w_1} \star \overrightarrow{w_2} \star \cdots \star
\overrightarrow{w_n}$, where $\star$ is a vector operator, such as
addition $+$, point-wise multiplication $\odot$, tensor product
$\otimes$, or matrix multiplication $\times$.

In the simplest compositional models (the first approach described
above), $\star$ is $+$ or $\odot$, e.g.~see
\cite{mitchell2008vector}. Grammar-based compositional models (the
third approach) are based on a generalisation of the notion of
vectors, known as \emph{tensors}. Whereas a vector $\ov{v}$ is an
element of an atomic vector space $V$, a tensor $\overline{z}$ is an
element of a tensor space $V \otimes W \otimes \cdots \otimes Z$. The
number of tensored spaces is referred to by the \emph{order} of the
space. Using a general duality theorem from multi-linear algebra
\cite{bourbaki}, it follows that tensors are in one-one correspondence
with multi-linear maps, that is we have:
\[
\overline{z} \in V \otimes W \otimes \cdots \otimes Z \  \cong \ f_{\overline{z}} \colon V \to W \to \cdots \to Z
\]

In such a tensor-based formalism, meanings of nouns are vectors and meanings of
predicates such as adjectives and verbs are tensors.  Meaning of a string of
words is obtained by applying the compositions of multi-linear map duals of the
tensors to the vectors.  For the sake of demonstration, take the case of an
intransitive sentence ``Sbj Verb''; the meaning of the subject is a vector
$\ov{\text{Sbj}} \in V$ and the meaning of the intransitive verb is a tensor
$\overline{\text{Verb}} \in V \otimes W$.  Meaning of the sentence is
obtained by applying $f_{\overline{Verb}}$ to $\ov{\text{Sbj}}$, as follows:
\[
\ov{\mbox{Sbj Verb}} = f_{\overline{Verb}} (\ov{\text{Sbj}})
\]

By tensor-map duality, the above becomes equivalent to the following, where composition has now become
the familiar notion of matrix multiplication, that is  $\star$ is $\times$:
\[
\overline{\text{Verb}} \times \ov{\text{Sbj}}
\]

In general and
for words with tensors of order higher than two, $\star$ becomes a
generalisation of $\times$, referred to by \emph{tensor contraction},
see e.g.~\newcite{KartsaklisEMNLP}. Since the creation and manipulation of tensors of order higher than 2 is difficult, 
one can work with simplified versions of tensors, faithful to their underlying mathematical basis; these have  found intuitive interpretations, e.g. see \newcite{grefenstette2011experimental},  \newcite{KartSadrQPL}. 
In such cases, $\star$ becomes a combination of a range of operations such as $\times$, $\otimes$, $\odot$,  and $+$.


\input{table-comp-methods.tex}

\paragraph{Specific models}

In the current paper we will experiment with a variety of models.  In
Table~\ref{tbl:comp-methods}, we present these models in terms of their composition operators 
and a reference to the main paper in which each
model was introduced. For the simple compositional models the sentence is a
string of any number of words; for the grammar-based models, we consider simple
transitive sentences ``$\mbox{Sbj Verb Obj}$'' and introduce the following
abbreviations for the concrete method used to build a tensor for the verb:

\begin{enumerate}
\item $\overline{\text{Verb}}$ is a verb matrix computed using the formula
  $\sum_i \overrightarrow{\text{Sbj}_i} \otimes \overrightarrow{\text{Obj}_i}$,
  where $\overrightarrow{\text{Sbj}_i}$ and $\overrightarrow{\text{Obj}_i}$ are
  the subjects and objects of the verb across the corpus. These models are
  referred to by \emph{relational} \cite{grefenstette2011experimental}; they are
  generalisations of predicate semantics of transitive verbs, from pairs of
  individuals to pairs of vectors. The models reduce
  the order 3 tensor of a transitive verb to an order 2 tensor (i.e.~a matrix).
\item $\widetilde{\text{Verb}}$ is a verb matrix computed using the formula
  $\overrightarrow{\text{Verb}} \otimes \overrightarrow{\text{Verb}}$, where
  $\overrightarrow{\text{Verb}}$ is the distributional vector of the verb. These
  models are referred to by \emph{Kronecker}, which is the term sometimes used
  to denote the outer product of tensors
  \cite{grefenstette2011gems}. This models also reduces
  the order 3 tensor of a transitive verb to an order 2 tensor.
\item The models of the last five lines of the table use the so-called
  \emph{Frobenius} operators from categorical compositional distributional
  semantics \cite{KartSadrCOLING} to expand the relational matrices of verbs from
  order 2 to order 3. The expansion is obtained by either copying the dimension
  of the subject into the space provided by the third tensor, hence referred to
  by \emph{Copy-Sbj}, or copying the dimension of the object in that space,
  hence referred to by \emph{Copy-Obj}; furthermore, we can take addition, multiplication, or
outer product of these, which are referred to by \emph{Frobenius-Add},
  \emph{Frobenius-Mult}, and \emph{Frobenius-Outer} \cite{KartSadrQPL}.
\end{enumerate}

\section{Semantic word spaces}
\label{sec:semantic-spaces}

Co-occurrence-based vector space instantiations have received a lot of
attention from the scientific community (refer to
\cite{kiela-clark:2014:CVSC,polajnar-clark:2014:EACL} for recent
studies). We instantiate two co-occurrence-based vectors spaces with
different underlying corpora and weighting schemes.

\paragraph{GS11}
\label{sec:ppmi}

Our first word space is based on a typical configuration that has been used in the past extensively for compositional distributional models (see below for details), so it will serve as a useful baseline for the current work. 
In this vector space, the co-occurrence counts are extracted from the British National Corpus (BNC)
\cite{leech1994claws4}. As basis words, we use the most frequent nouns, verbs,
adjectives and adverbs (POS tags \texttt{SUBST}, \texttt{VERB},
\texttt{ADJ} and \texttt{ADV} in the BNC XML
distribution\footnote{\url{http://www.natcorp.ox.ac.uk/}}). 
The vector space is lemmatized, that is, it contains only ``canonical'' forms of words.

In order to weight the raw co-occurrence counts, we use positive point-wise mutual information (PPMI). 
The component value for a target word $t$ and a context word $c$ is given by:
\begin{equation*}
  \operatorname{PPMI}(t, c)= \max\left(
    0,
    \log 
       \frac{p(c|t)}{p(c)}
  \right)
\end{equation*}
\noindent
where $p(c|t)$ is the probability of word $c$ given $t$ in a symmetric window of length 5 and $p(c)$ is the probability of $c$ 
overall.

Vector spaces based on point-wise mutual information (or variants
thereof) have been successfully applied in various distributional and
compositional tasks; see e.g. \newcite{grefenstette2011experimental}, \newcite{mitchell2008vector}, \newcite{levy2014linguistic}
for details. PPMI has been shown to achieve state-of-the-art results
\cite{levy2014linguistic} and is suggested by the review of
\newcite{kiela-clark:2014:CVSC}. Our use here of the BNC as a corpus
and the window length of 5 is based on previous use and better performance of these
parameters in a number of compositional experiments
\cite{grefenstette2011experimental,grefenstette2011gems,mitchell2008vector,KartSadrCOLING}.

\paragraph{KS14}

In this variation, we train a vector space from the ukWaC
corpus\footnote{\url{http://wacky.sslmit.unibo.it/}} \cite{ukwac}, originally
using as a basis the 2,000 content words with the highest frequency (but
excluding a list of stop words as well as the 50 most frequent content words
since they exhibit low information content). The vector space is again lemmatized. As context we consider a 5-word
window from either side of the target word, while as our weighting scheme we use
local mutual information (i.e.~point-wise mutual information multiplied by raw
counts). In a further step, the vector space was normalized and projected onto a 300-dimensional space
using singular value decomposition (SVD).


In general, dimensionality reduction produces more compact word
representations that are robust against potential noise in the corpus \cite{Landauer,schutze1997ambiguity}. SVD
has been shown to perform well on a variety of tasks similar to ours \cite{baroni2010nouns,KartSadrQPL}.

\paragraph{Neural word embeddings (NWE)}
\label{sec:neur-word-embedd}

For our neural setting, we used the skip-gram model of \newcite{mikolov2013distributed} trained with negative sampling. 
The specific implementation that was tested in our experiments was a 300-dimensional vector space learned from the Google News corpus and provided by the \texttt{word2vec}\footnote{\url{https://code.google.com/p/word2vec/}} toolkit. 
Furthermore, the \texttt{gensim} library \cite{rehurek_lrec} was used for accessing the vectors. 
On the contrary with the previously described co-occurrence vector spaces, this version is \textit{not} lemmatized.

The negative sampling method improves the objective function of Equation \ref{eq:objective-func} by introducing negative examples to the training algorithm. 
Assume that the probability of a specific $(c,t)$ pair of words (where $t$ is a target word and $c$ another word in the same context with $t$), 
coming from the training data, is denoted as $p(D=1|c,t)$. The objective function is then expressed as follows:
\begin{equation}
\label{eq:neg1}
\prod\limits_{(c,t)\in D}p(D=1|c,t)
\end{equation}
\noindent
That is, the goal is to set the model parameters in a way that maximizes the probability of all observations coming from the training data. 
Assume now that $D'$ is a set of randomly selected incorrect $(c',t')$ pairs that do not occur in $D$, then Equation \ref{eq:neg1} above can be recasted in the following way:
\begin{equation}
\label{eq:neg2}
\prod\limits_{(c,t)\in D}p(D=1|c,t) \prod\limits_{(c',t')\in D'}p(D=0|c',t')
\end{equation}
In other words, the model tries to distinguish a target word $t$ from random draws that come from a noise distribution. 
In the implementation we used for our experiments, $c$ is always selected from a 5-word window around $t$. 
More details about the negative sampling approach can be found in \cite{mikolov2013distributed}; 
the note of \newcite{goldberg2014word2vec} also provides an intuitive explanation of the underlying setting.


\section{Experiments}
\label{sec:experiments}

Our experiments explore the use of the vector spaces above, together with the compositional operators described in Section~\ref{sec:compositional-models}, 
in a range of tasks all of
which require semantic composition: verb sense disambiguation;
sentence similarity; paraphrasing; and dialogue act tagging.


\subsection{Disambiguation}
\label{sec:disamb}

We use the transitive verb disambiguation dataset described
in \newcite{grefenstette2011experimental}\footnote{This and
  the sentence similarity dataset are available at
  \url{http://www.cs.ox.ac.uk/activities/compdistmeaning/}}. This dataset consists of ambiguous
transitive verbs together with their arguments,  landmark verbs that
identify one of the verb senses, 
and human judgements that specify how similar is the disambiguated sense of the verb in the given context to one of the landmarks. 
This is similar to the intransitive dataset
described in \cite{mitchell2008vector}. Consider the sentence ``system meets specification''; here,
\textit{meets} is the ambiguous transitive verb, and \textit{system}
and \textit{specification} are its arguments in this context. Possible
landmarks for \emph{meet} are \textit{satisfy} and \textit{visit}; for
this sentence, the human judgements show that the disambiguated 
meaning of the verb is more similar to the landmark \textit{satisfy} and less similar
to \textit{visit}.

The task is to estimate the similarity of the sense of a verb in a context with
a given landmark. To get our similarity measures, we compose the verb with its
arguments using one of our compositional models; we do the same for
the landmark and then compute the cosine similarity of
the two vectors. We evaluate the performance by averaging the human
judgements for the same verb, argument and landmark entries, and calculating the Spearman's correlation  between
the average values and the cosine scores. As a baseline, we
compare this with the correlation produced by using only the verb vector,
without composing it with its arguments.

\input{table-wsd-results.tex}

Table~\ref{tab:wsd-results} shows the results of the experiment. NWE
\textit{copy-object} composition yields the best correlation with the human
judgements, and top performance across all vector spaces and models with a Spearman~$\rho$ of 0.456. 
For the KS14 space, the best result comes from \textit{Frobenius outer} (0.350), 
while the best operator for the GS11 space is \textit{point-wise multiplication} (0.348).

For simple point-wise composition, only multiplicative GS11 and additive NWE
improve over their corresponding verb-only baselines (but both perform
worse than the KS14 baseline). With tensor-based composition in co-occurrence
based spaces, \textit{copy subject} yields lower results than the corresponding
baselines. Other composition methods, except \textit{Kronecker} for
KS14, improve over the verb-only baselines. 
Finally we should note that, despite the small training corpus, the GS11 vector space performs comparatively well:
for instance, \textit{Kronecker} model improves the previously reported score of 0.28 \cite{grefenstette2011gems}.

\subsection{Sentence similarity}
\label{sec:sentence-similarity}

In this experiment we use the transitive sentence similarity dataset
described in \newcite{KartSadrQPL}. The dataset consists of
transitive sentence pairs and a human similarity judgement\footnote{The textual content of this dataset is the same as that of \cite{KartsaklisEMNLP},  the difference is that the dataset of \cite{KartSadrQPL}  has updated  human judgements whereas the previous dataset used  the original annotations of the  intransitive dataset of \cite{lapata2010}.}. The task
is to estimate a  similarity measure between two sentences. As in the
disambiguation task, we first compose word vectors to obtain sentence
vectors, then compute cosine similarity of them. We average the human
judgements for identical sentence pairs to compute a correlation with cosine scores.

\input{table-sent-sim-results.tex}

Table~\ref{tab:sent-sim-results} shows the results. Again, the best performing vector space is KS14, but this
time with \textit{addition}: the Spearman~$\rho$ correlation score
with averaged human judgements is 0.732. Addition was the means for the other vector spaces to achieve top
performance as well: GS11 and NWE got 0.682 and 0.689 respectively.

None of the models in tensor-based composition outperformed
addition. KS14 performs worse with tensor-based methods here than in 
the other vector spaces. However, GS11 and NWE, except \textit{copy subject}
for both of them and \textit{Frobenius multiplication} for NWE, improved over
their verb-only baselines.

\subsection{Paraphrasing}
\label{sec:paraphrasing}

In this experiment we evaluate our vector spaces on a mainstream
paraphrase detection task. Specifically, we get classification results
on the Microsoft Research Paraphrase Corpus paraphrase corpus
\cite{dolan2005microsoft} working in the following way: we construct
vectors for the sentences of each pair; if the cosine similarity
between the two sentence vectors exceeds a certain threshold, the pair
is classified as a paraphrase, otherwise as not a paraphrase.  For
this experiment and that of Section~\ref{sec:dialogue-act-tagging}
below, we investigate only the addition and  point-wise
multiplication compositional models, 
since at their current stage of development tensor-based models can only efficiently handle sentences of fixed structure. 
Nevertheless, the simple point-wise compositional models still allow for a direct comparison of the vector spaces, 
which is the main goal of this paper.

For each vector space and model, a number of different thresholds were tested on
the first 2000 pairs of the training set, which we used as a development set; in
each case, the best-performed threshold was selected for a \textit{single} run
of our ``classifier'' on the test set (1726 pairs). Additionally, we evaluate
the NWE model with a lemmatized version of the corpus, so that the experimental setup
is maximally similar for all vector spaces.  The results are shown in the first
part of Table~\ref{tbl:mspr}.

\input{table-par-results.tex}

Additive NWE gives the highest performance, with both lemmatized and
un-lemmatized versions outperforming the GS11 and KS14 spaces. In the
un-lemmatized case, the accuracy of our simple ``classifier'' (0.73) is close to
state-of-the-art range. The state-of-the art result (0.77 accuracy and 0.84
F-score\footnote{F-scores use the standard definition $F =
  2(\mathit{precision} * \mathit{recall}) / (\mathit{precision} + \mathit{recall})$.})
by the time of this writing has been obtained using 8 machine translation metrics and three
constituent classifiers \cite{madnani2012re}.

The multiplicative model gives lower results than the additive model across all
vector spaces. The KS14 vector space shows the steadiest
performance, with a drop in accuracy of only 0.04 and no drop in
F-score, while for the GS11 and NWE spaces both accuracy and F-score
experienced drops by more than 0.20.

\subsection{Dialogue act tagging}
\label{sec:dialogue-act-tagging}

As our last experiment, we evaluate the word spaces on a dialogue act tagging task
\cite{Stolcke.etal00} over the Switchboard corpus \cite{godfrey1992switchboard}.  Switchboard is a collection of
approximately 2500 dialogs over a telephone line by 500 speakers from
the U.S. on predefined topics.\footnote{The dataset and a Python
  interface to it are available at
  \url{http://compprag.christopherpotts.net/swda.html}}

The experiment pipeline follows \cite{milajevs-purver:2014:CVSC}. The
input utterances are preprocessed so that the parts of interrupted
utterances are concatenated \cite{webb2005dialogue}. Disfluency markers and commas are removed
from the utterance raw texts. For GS11 and KS14 the
utterance tokens are POS-tagged and lemmatized; for NWE, we
test the vectors in both a lemmatized and an un-lemmatized version of the corpus.\footnote{We use \texttt{WordNetLemmatizer} of
  the NLTK library \cite{bird2006nltk}.} We split the training and testing
utterances as suggested by \newcite{Stolcke.etal00}.
Utterance vectors are then obtained as in the previous experiments;
they are reduced to 50 dimensions using SVD and a $k$-nearest-neighbour
classifier is trained on these reduced utterance vectors (the 5 closest
neighbours by Euclidean distance are retrieved to make a
classification decision). The results are shown in the second part of
Table~\ref{tbl:mspr}.

Un-lemmatized NWE \textit{addition} gave the best accuracy (0.63) and
F-score (0.60) (averaged over tag classes), i.e.~similar results to \cite{milajevs-purver:2014:CVSC}---although note that the dimensionality of our NWE vectors is 10 times lower than theirs. 
\textit{Multiplicative} NWE outperformed the corresponding model in \cite{milajevs-purver:2014:CVSC}. 
In general, addition consistently outperforms multiplication for all
the models.  Lemmatization dramatically lowers tagging accuracy: the lemmatized GS11, KS14 and NWE models perform much worse than un-lemmatized NWE, suggesting that morphological features are important for this task.


\section{Discussion}
\label{sec:discussion}

Previous comparisons of co-occurrence-based and neural word vector
representations vary widely in their conclusions. While
\newcite{baroni2014don} conclude that ``context-predicting
  models obtain a thorough and resounding victory against their
  count-based counterparts'', this seems to contradict, at least at
the first consideration, the more conservative conclusion of
\newcite{levy2014linguistic} that ``analogy recovery is not
  restricted to neural word embeddings [\ldots] a similar amount of
  relational similarities can be recovered from traditional
  distributional word representations'' and the findings of
\newcite{blacoe2012comparison} that ``shallow approaches are
  as good as more computationally intensive alternatives'' on phrase
similarity and paraphrase detection tasks.

It seems clear that neural word embeddings have an advantage when used
in tasks for which they have been trained; our main questions here are
whether they outperform co-occurrence based alternatives across the
board; and which approach lends itself better to composition using
general mathematical operators. To partially answer this question, we
can compare model behaviour against the baselines in \textit{isolation}.

For the disambiguation and sentence similarity tasks the baseline is
the similarity between verbs only, ignoring the context---see
above. For the paraphrase task, we take the global vector-based
similarity reported in \cite{mihalcea2006corpus}: 0.65 accuracy and
0.75 F-score. For the dialogue act tagging task the baseline is the
accuracy of the bag-of-unigrams model in
\cite{milajevs-purver:2014:CVSC}: 0.60.

Sections~\ref{sec:disamb} and \ref{sec:sentence-similarity} show that although the best choice of vector representation might vary, for small-scale tasks all methods give fairly competitive results. 
The choice of compositional operator seems to be more important and more task-specific: 
while a tensor-based operation (Frobenius copy-object) performs best for verb disambiguation, the best result for sentence similarity is achieved by a simple additive model, with all other compositional methods behaving worse than the verb-only baseline in the KS14 case. 
GS11 and NWE, on the other hand, outperform their baselines with a number of compositional methods, although both of 
them achieve lower performance than KS14 overall.

Based on only small-scale experiment results, one could conclude that there is little significant difference between 
the two ways of obtaining vectors. GS11 and NWE show similar behaviour in comparison to their baselines, 
while it is possible to tune a co-occurrence based vector space (KS14) and obtain the best result. 
Large scale tasks reveal another pattern: the GS11 vector space, which behaves stably on the small scale, 
drags behind the KS14 and NWE spaces in the paraphrase detection task. 
In addition, NWE consistently yields best results. Finally, only the NWE space was able to provide adequate results on the dialogue act tagging task. Table~\ref{tab:summary} 
summarizes model performance with regard to baselines.

\input{table-summary.tex}

\section{Conclusion}
\label{sec:concl-future-work}

In this work we compared the performance of two co-occurrence-based semantic
spaces with vectors learned by a neural network in compositional
settings. We carried out two small-scale tasks (word sense disambiguation and
sentence similarity) and two large-scale tasks (paraphrase detection and
dialogue act tagging).

On small-scale tasks, where the sentence structures are predefined and
relatively constrained, NWE gives better or similar
results to count-based vectors. Tensor-based composition does not always outperform simple
compositional operators, but for most of the cases gives results within the same range.

On large-scale tasks, neural vectors are more
successful than the co-occurrence based alternatives. However, this
study does not reveal whether this is because of their neural nature, or
just because they are trained on a larger amount of data.

The question of whether neural vectors outperform co-occurrence
vectors therefore requires further detailed comparison to be entirely resolved; 
our experiments suggest that this is indeed the case in
large-scale tasks, but the difference in size and nature of the
original corpora may be a confounding factor.  In any case, it is clear
that the neural vectors of \texttt{word2vec} package perform steadily off-the-shelf
across a large variety of tasks. The size of the vector space (3 million words) and
the available code-base that simplifies the access to the vectors, makes this set a good and safe choice 
for experiments in the future. Of course, even better performances can 
be achieved by training neural language models specifically for a given task (see e.g.~\newcite{KalchbrennerACL2014}).

The choice of compositional operator (tensor-based or a simple point-wise operation) depends strongly on the task and dataset: 
tensor-based composition performed best with the verb disambiguation task, 
where the verb senses depend strongly on the arguments of the verb. 
However, it seems to depend less on the nature of the vectors itself: 
in the disambiguation task, tensor-based composition proved best for both co-occurrence-based and neural vectors; 
in the sentence similarity task, where point-wise operators proved best, this was again true across vector spaces.



\section*{Acknowledgements}

We would like to thank the three anonymous reviewers for their fruitful comments. 
Support by EPSRC grant EP/F042728/1 is gratefully acknowledged by Milajevs, Kartsaklis and Sadrzadeh. 
Purver is partly supported by ConCreTe: the project ConCreTe
acknowledges the financial support of the Future and Emerging
Technologies (FET) programme within the Seventh Framework Programme
for Research of the European Commission, under FET grant number
611733.

\bibliographystyle{acl}
\bibliography{references}

\end{document}

%% file: table-comp-methods.tex
\begin{table*}[hbt]
  \begin{center}
    \footnotesize
    \begin{tabular}{llll}
      \toprule
      \textbf{Method} & \textbf{Sentence} & \textbf{Linear algebraic formula} & \textbf{Reference}\\
      \midrule
      Addition &
      $w_1 w_2 \cdots w_n$ &
      $\overrightarrow{w_1} + \overrightarrow{w_2} + \cdots + \overrightarrow{w_n}$ &
      \newcite{mitchell2008vector}
      \\
      Multiplication &
      $w_1 w_2 \cdots w_n$ &
      $\overrightarrow{w_1} \odot \overrightarrow{w_2} \odot \cdots \odot \overrightarrow{w_n}$ &
      \newcite{mitchell2008vector}
      \\
      \midrule
      Relational &
      $\mbox{Sbj Verb Obj}$ &
      $\overline{\text{Verb}} \odot (\overrightarrow{\text{Sbj}} \otimes \overrightarrow{\text{Obj}})$ &
      \newcite{grefenstette2011experimental}
      \\
      Kronecker &
      $\mbox{Sbj Verb Obj}$ & $\widetilde{\text{Verb}} \odot (\overrightarrow{\text{Sbj}} \otimes \overrightarrow{\text{Obj}})$ &
      \newcite{grefenstette2011gems}
      \\
      \midrule
      Copy object&
       $\mbox{Sbj Verb Obj}$ & $\overrightarrow{\text{Sbj}} \odot (\overline{\text{Verb}} \times \overrightarrow{\text{Obj}})$ &
      \newcite{KartSadrCOLING}
      \\
      Copy subject&
      $\mbox{Sbj Verb Obj}$& $\overrightarrow{\text{Obj}} \odot (\overline{\text{Verb}}^{\mathsf{T}} \times \overrightarrow{\text{Sbj}})$ &
      \newcite{KartSadrCOLING}
      \\
      Frob. add.&
       $\mbox{Sbj Verb Obj}$ & $(\overrightarrow{\text{Sbj}} \odot (\overline{\text{Verb}} \times \overrightarrow{\text{Obj}})) +
      (\overrightarrow{\text{Obj}} \odot (\overline{\text{Verb}}^{\mathsf{T}} \times \overrightarrow{\text{Sbj}}))$ &
      \newcite{KartSadrQPL}
      \\
      Frob. mult.&
      $\mbox{Sbj Verb Obj}$ &
      $(\overrightarrow{\text{Sbj}} \odot (\overline{\text{Verb}} \times \overrightarrow{\text{Obj}})) \odot
      (\overrightarrow{\text{Obj}} \odot (\overline{\text{Verb}}^{\mathsf{T}} \times \overrightarrow{\text{Sbj}}))$ &
      \newcite{KartSadrQPL}
      \\
      Frob. outer&
     $\mbox{Sbj Verb Obj}$ &
      $(\overrightarrow{\text{Sbj}} \odot (\overline{\text{Verb}} \times \overrightarrow{\text{Obj}})) \otimes
      (\overrightarrow{\text{Obj}} \odot (\overline{\text{Verb}}^{\mathsf{T}} \times \overrightarrow{\text{Sbj}}))$ &
      \newcite{KartSadrQPL}
      \\
      \bottomrule
    \end{tabular}
    \caption{Compositional methods.}
    \label{tbl:comp-methods}
  \end{center}
\end{table*}


%% file: table-wsd-results.tex
\begin{table}[b!]
  \centering
  \begin{tabular}{lrrr}
    \toprule
    \textbf{Method} &
    \textbf{GS11} &
    \textbf{KS14} &
    \textbf{NWE}
    \\
    \midrule
    Verb only &
    0.212 &
    0.325 &
    0.107
    \\
    \midrule
    Addition &
    0.103 &
    0.275 &
    0.149
    \\
    Multiplication &
    0.348 &
    0.041 &
    0.095
    \\
    \midrule
    Kronecker &
    0.304 &
    0.176 &
    0.117
    \\
    Relational &
    0.285 &
    0.341 &
    0.362
    \\
    Copy subject &
    0.089 &
    0.317 &
    0.131
    \\
    Copy object &
    0.334 &
    0.331 &
    \textbf{0.456}
    \\
    Frobenius add. &
    0.261 &
    0.344 &
    0.359
    \\
    Frobenius mult. &
    0.233 &
    0.341 &
    0.239
    \\
    Frobenius outer &
    0.284 &
    0.350 &
    0.375
    \\
    \bottomrule
  \end{tabular}
  \caption{Spearman~$\rho$ correlations of models with human judgements
    for the word sense disambiguation task. The best result (NWE Copy
    object) outperforms the nearest co-occurrence-based competitor (KS14 Frobenius outer) with a statistically significant difference ($p < 0.05$, t-test).}
  \label{tab:wsd-results}
\end{table}


%% file: table-sent-sim-results.tex
\begin{table}[hbt]
  \centering
  \begin{tabular}{lrrr}
    \toprule
    \textbf{Method} &
    \textbf{GS11} &
    \textbf{KS14} &
    \textbf{NWE}
    \\
    \midrule
    Verb only &
    0.491 &
    0.602 &
    0.561
    \\
    \midrule
    Addition &
    0.682 &
    \textbf{0.732} &
    0.689
    \\
    Multiplication &
    0.597 &
    0.321 &
    0.341
    \\
    \midrule
    Kronecker &
    0.581 &
    0.408 &
    0.561
    \\
    Relational &
    0.558 &
    0.437 &
    0.618
    \\
    Copy subject &
    0.370 &
    0.448 &
    0.405
    \\
    Copy object &
    0.571 &
    0.306 &
    0.655
    \\
    Frobenius add. &
    0.566 &
    0.460 &
    0.585
    \\
    Frobenius mult. &
    0.525 &
    0.226 &
    0.387
    \\
    Frobenius outer &
    0.560 &
    0.439 &
    0.622
    \\
    \bottomrule
  \end{tabular}
  \caption{Results for sentence similarity. There is no statistically
    significant difference between KS14 addition and NWE addition (the second
    best result).}
  \label{tab:sent-sim-results}
\end{table}


%% file: table-par-results.tex
\begin{table*}[hbt]
  \begin{center}
    \scriptsize
    \begin{tabular}{lcccccccccc}
      \toprule
      &
      &
      &
      \multicolumn{4}{c}{\textbf{Co-occurrence}} &
      \multicolumn{4}{c}{\textbf{Neural word embeddings}}
      \\
      \cmidrule(r){4-7}
      \cmidrule(r){8-11}
      &
      \multicolumn{2}{c}{\textbf{Baseline}} &
      \multicolumn{2}{c}{\textbf{GS11}} &
      \multicolumn{2}{c}{\textbf{KS14}} &
      \multicolumn{2}{c}{\textbf{Unlemmatized}} &
      \multicolumn{2}{c}{\textbf{Lemmatized}}
      \\
      \cmidrule(r){2-3}
      \cmidrule(r){4-5}
      \cmidrule(r){6-7}
      \cmidrule(r){8-9}
      \cmidrule(r){10-11}
      Model &
      Accuracy & F-Score &
      Accuracy & F-Score &
      Accuracy & F-Score &
      Accuracy & F-Score &
      Accuracy & F-Score
      \\
      \midrule
      MSR addition &
      \multirow{2}{*}{0.65} & \multirow{2}{*}{0.75} &
      0.62 & 0.79 &
      0.70 & 0.80 &
      \textbf{0.73} & \textbf{0.82} &
      0.72 & 0.81
      \\
      MSR multiplication &
      & &
      0.52 & 0.58 &
      \textbf{0.66} & \textbf{0.80} &
      0.42 & 0.34 &
      0.41 & 0.36
      \\
     \midrule
      SWDA addition &
      \multirow{2}{*}{0.60} & \multirow{2}{*}{0.58} &
      0.35 & 0.35 &
      0.40 & 0.35 &
      \textbf{0.63} & \textbf{0.60} &
      0.44 & 0.40
      \\
      SWDA multiplication &
      & &
      0.32 & 0.16 &
      0.39 & 0.33 &
      \textbf{0.58} & \textbf{0.53} &
      0.43 & 0.38
      \\
      \bottomrule
    \end{tabular}
    \caption{Results for paraphrase detection (MSR) and dialog act tagging
      (SWDA) tasks. All top results significantly outperform corresponding
      nearest competitors (for accuracy): $p < 0.05$, $\chi^2$ test.}
    \label{tbl:mspr}
  \end{center}
\end{table*}


%% file: table-summary.tex
\begin{table}[tbh]
  \centering
  \begin{tabular}{lccc}
    \toprule
    \textbf{Task} &
    \textbf{GS11} &
    \textbf{KS14} &
    \textbf{NWE}
    \\
    \midrule
    Disambiguation &
    $+$ &
    $+$ &
    $\textbf{\large+}$
    \\
    Sentence similarity &
    $+$ &
    $\textbf{\large--}$ &
    $+$
    \\
    \midrule
    Paraphrase &
    $-$ &
    $+$ &
    $\textbf{\large+}$
    \\
    Dialog act tagging &
    $-$ &
    $-$ &
    $\textbf{\large+}$
    \\
    \bottomrule
  \end{tabular}
  \caption{Summary of vector space performance against baselines. General
    improvement (cases where more than a half of the models perform better)
    and decrease with regard to a corresponding baseline is respectively marked
    by $+$ and $-$. A bold value means that the model gave the best result in the task.}
  \label{tab:summary}
\end{table}


%% file: main.bbl
\begin{thebibliography}{}

\bibitem[\protect\citename{Baroni and Zamparelli}2010]{baroni2010nouns}
Marco Baroni and Roberto Zamparelli.
\newblock 2010.
\newblock Nouns are vectors, adjectives are matrices: Representing
  adjective-noun constructions in semantic space.
\newblock In {\em Proceedings of the 2010 Conference on Empirical Methods in
  Natural Language Processing}, pages 1183--1193. Association for Computational
  Linguistics.

\bibitem[\protect\citename{Baroni \bgroup et al.\egroup }2014]{baroni2014don}
Marco Baroni, Georgiana Dinu, and Germ{\'a}n Kruszewski.
\newblock 2014.
\newblock Don't count, predict! a systematic comparison of context-counting vs.
  context-predicting semantic vectors.
\newblock In {\em Proceedings of the 52nd Annual Meeting of the Association for
  Computational Linguistics}, volume~1.

\bibitem[\protect\citename{Bengio \bgroup et al.\egroup }2006]{bengio2006}
Yoshua Bengio, Holger Schwenk, Jean-S{\'e}bastien Sen{\'e}cal, Fr{\'e}deric
  Morin, and Jean-Luc Gauvain.
\newblock 2006.
\newblock Neural probabilistic language models.
\newblock In {\em Innovations in Machine Learning}, pages 137--186. Springer.

\bibitem[\protect\citename{Bird}2006]{bird2006nltk}
Steven Bird.
\newblock 2006.
\newblock {NLTK: the natural language toolkit}.
\newblock In {\em Proceedings of the COLING/ACL on Interactive presentation
  sessions}, pages 69--72. Association for Computational Linguistics.

\bibitem[\protect\citename{Blacoe and Lapata}2012]{blacoe2012comparison}
William Blacoe and Mirella Lapata.
\newblock 2012.
\newblock A comparison of vector-based representations for semantic
  composition.
\newblock In {\em Proceedings of the 2012 Joint Conference on Empirical Methods
  in Natural Language Processing and Computational Natural Language Learning},
  pages 546--556. Association for Computational Linguistics.

\bibitem[\protect\citename{Bos and Gabsdil}2000]{bos2000first}
Johan Bos and Malte Gabsdil.
\newblock 2000.
\newblock First-order inference and the interpretation of questions and
  answers.
\newblock {\em Proceedings of Gotelog}, pages 43--50.

\bibitem[\protect\citename{Bos}2008]{Bos2008STEP2}
Johan Bos.
\newblock 2008.
\newblock Wide-coverage semantic analysis with boxer.
\newblock In Johan Bos and Rodolfo Delmonte, editors, {\em Semantics in Text
  Processing. STEP 2008 Conference Proceedings}, Research in Computational
  Semantics, pages 277--286. College Publications.

\bibitem[\protect\citename{Bourbaki}1989]{bourbaki}
N.~Bourbaki.
\newblock 1989.
\newblock {\em Commutative Algebra: Chapters 1-7}.
\newblock Srpinger Verlag, Berlin/New York.

\bibitem[\protect\citename{Coecke \bgroup et al.\egroup }2010]{coecke2010}
Bob Coecke, Mehrnoosh Sadrzadeh, and Stephen Clark.
\newblock 2010.
\newblock Mathematical foundations for a compositional distributional model of
  meaning.
\newblock {\em CoRR}, abs/1003.4394.

\bibitem[\protect\citename{Collobert and Weston}2008]{collobert2008}
Ronan Collobert and Jason Weston.
\newblock 2008.
\newblock A unified architecture for natural language processing: Deep neural
  networks with multitask learning.
\newblock In {\em Proceedings of the 25th international conference on Machine
  learning}, pages 160--167. ACM.

\bibitem[\protect\citename{Copestake \bgroup et al.\egroup
  }2005]{copestake2005minimal}
Ann Copestake, Dan Flickinger, Carl Pollard, and Ivan~A Sag.
\newblock 2005.
\newblock Minimal recursion semantics: An introduction.
\newblock {\em Research on Language and Computation}, 3(2-3):281--332.

\bibitem[\protect\citename{Dolan \bgroup et al.\egroup
  }2005]{dolan2005microsoft}
Bill Dolan, Chris Brockett, and Chris Quirk.
\newblock 2005.
\newblock Microsoft research paraphrase corpus.
\newblock {\em Retrieved May}, 29:2013.

\bibitem[\protect\citename{Ferraresi \bgroup et al.\egroup }2008]{ukwac}
Adriano Ferraresi, Eros Zanchetta, Marco Baroni, and Silvia Bernardini.
\newblock 2008.
\newblock Introducing and evaluating uk{W}a{C}, a very large web-derived corpus
  of {E}nglish.
\newblock In {\em Proceedings of the 4th Web as Corpus Workshop (WAC-4) Can we
  beat Google}, pages 47--54.

\bibitem[\protect\citename{Frege}1892]{frege1892sense}
Gottlob Frege.
\newblock 1892.
\newblock On sense and reference.
\newblock {\em Ludlow (1997)}, pages 563--584.

\bibitem[\protect\citename{Godfrey \bgroup et al.\egroup
  }1992]{godfrey1992switchboard}
John~J Godfrey, Edward~C Holliman, and Jane McDaniel.
\newblock 1992.
\newblock Switchboard: Telephone speech corpus for research and development.
\newblock In {\em Acoustics, Speech, and Signal Processing, 1992. ICASSP-92.,
  1992 IEEE International Conference on}, volume~1, pages 517--520. IEEE.

\bibitem[\protect\citename{Goldberg and Levy}2014]{goldberg2014word2vec}
Yoav Goldberg and Omer Levy.
\newblock 2014.
\newblock word2vec {E}xplained: deriving {M}ikolov et al.'s negative-sampling
  word-embedding method.
\newblock {\em arXiv preprint arXiv:1402.3722}.

\bibitem[\protect\citename{Grefenstette and
  Sadrzadeh}2011a]{grefenstette2011experimental}
Edward Grefenstette and Mehrnoosh Sadrzadeh.
\newblock 2011a.
\newblock Experimental support for a categorical compositional distributional
  model of meaning.
\newblock In {\em Proceedings of the Conference on Empirical Methods in Natural
  Language Processing}, pages 1394--1404. Association for Computational
  Linguistics.

\bibitem[\protect\citename{Grefenstette and
  Sadrzadeh}2011b]{grefenstette2011gems}
Edward Grefenstette and Mehrnoosh Sadrzadeh.
\newblock 2011b.
\newblock Experimenting with transitive verbs in a {DisCoCat}.
\newblock In {\em Proceedings of the GEMS 2011 Workshop on GEometrical Models
  of Natural Language Semantics}, pages 62--66, Edinburgh, UK, July.
  Association for Computational Linguistics.

\bibitem[\protect\citename{Harris}1954]{harris1954distributional}
Z.S. Harris.
\newblock 1954.
\newblock Distributional structure.
\newblock {\em Word}.

\bibitem[\protect\citename{Kalchbrenner and
  Blunsom}2013]{kalchbrenner-blunsom2013CVSC}
Nal Kalchbrenner and Phil Blunsom.
\newblock 2013.
\newblock Recurrent convolutional neural networks for discourse
  compositionality.
\newblock In {\em Proceedings of the Workshop on Continuous Vector Space Models
  and their Compositionality}, pages 119--126, Sofia, Bulgaria, August.
  Association for Computational Linguistics.

\bibitem[\protect\citename{Kalchbrenner \bgroup et al.\egroup
  }2014]{KalchbrennerACL2014}
Nal Kalchbrenner, Edward Grefenstette, and Phil Blunsom.
\newblock 2014.
\newblock A convolutional neural network for modelling sentences.
\newblock {\em Proceedings of the 52nd Annual Meeting of the Association for
  Computational Linguistics}, June.

\bibitem[\protect\citename{Kartsaklis and Sadrzadeh}2013]{KartsaklisEMNLP}
Dimitri Kartsaklis and Mehrnoosh Sadrzadeh.
\newblock 2013.
\newblock Prior disambiguation of word tensors for constructing sentence
  vectors.
\newblock In {\em Proceedings of the 2013 Conference on Empirical Methods in
  Natural Language Processing (EMNL)}, pages 1590--1601, Seattle, USA, October.
  Association for Computational Linguistics.

\bibitem[\protect\citename{Kartsaklis and Sadrzadeh}2014]{KartSadrQPL}
Dimitri Kartsaklis and Mehrnoosh Sadrzadeh.
\newblock 2014.
\newblock A study of entanglement in a categorical framework of natural
  language.
\newblock In {\em Proceedings of the 11th Workshop on Quantum Physics and Logic
  (QPL)}, Kyoto, Japan, June.

\bibitem[\protect\citename{Kartsaklis \bgroup et al.\egroup
  }2012]{KartSadrCOLING}
Dimitri Kartsaklis, Mehrnoosh Sadrzadeh, and Stephen Pulman.
\newblock 2012.
\newblock A unified sentence space for categorical distributional-compositional
  semantics: Theory and experiments.
\newblock In {\em Proceedings of COLING 2012: Posters}, pages 549--558, Mumbai,
  India, December. The COLING 2012 Organizing Committee.

\bibitem[\protect\citename{Kiela and Clark}2014]{kiela-clark:2014:CVSC}
Douwe Kiela and Stephen Clark.
\newblock 2014.
\newblock A systematic study of semantic vector space model parameters.
\newblock In {\em Proceedings of the 2nd Workshop on Continuous Vector Space
  Models and their Compositionality (CVSC)}, pages 21--30, Gothenburg, Sweden,
  April. Association for Computational Linguistics.

\bibitem[\protect\citename{Landauer and Dumais}1997]{Landauer}
T.~Landauer and S.~Dumais.
\newblock 1997.
\newblock {A} {S}olution to {P}lato's {P}roblem: {T}he {L}atent {S}emantic
  {A}nalysis {T}heory of {A}cquision, {I}nduction, and {R}epresentation of
  {K}nowledge.
\newblock {\em Psychological Review}.

\bibitem[\protect\citename{Leech \bgroup et al.\egroup }1994]{leech1994claws4}
Geoffrey Leech, Roger Garside, and Michael Bryant.
\newblock 1994.
\newblock Claws4: the tagging of the british national corpus.
\newblock In {\em Proceedings of the 15th conference on Computational
  linguistics-Volume 1}, pages 622--628. Association for Computational
  Linguistics.

\bibitem[\protect\citename{Levy \bgroup et al.\egroup
  }2014]{levy2014linguistic}
Omer Levy, Yoav Goldberg, and Israel Ramat-Gan.
\newblock 2014.
\newblock Linguistic regularities in sparse and explicit word representations.
\newblock In {\em Proceedings of the Eighteenth Conference on Computational
  Natural Language Learning, Baltimore, Maryland, USA, June. Association for
  Computational Linguistics}.

\bibitem[\protect\citename{Madnani \bgroup et al.\egroup }2012]{madnani2012re}
Nitin Madnani, Joel Tetreault, and Martin Chodorow.
\newblock 2012.
\newblock Re-examining machine translation metrics for paraphrase
  identification.
\newblock In {\em Proceedings of the 2012 Conference of the North American
  Chapter of the Association for Computational Linguistics: Human Language
  Technologies}, pages 182--190. Association for Computational Linguistics.

\bibitem[\protect\citename{Mihalcea \bgroup et al.\egroup
  }2006]{mihalcea2006corpus}
Rada Mihalcea, Courtney Corley, and Carlo Strapparava.
\newblock 2006.
\newblock Corpus-based and knowledge-based measures of text semantic
  similarity.
\newblock In {\em AAAI}, volume~6, pages 775--780.

\bibitem[\protect\citename{Mikolov \bgroup et al.\egroup
  }2013a]{mikolov2013efficient}
Tomas Mikolov, Kai Chen, Greg Corrado, and Jeffrey Dean.
\newblock 2013a.
\newblock Efficient estimation of word representations in vector space.
\newblock {\em arXiv preprint arXiv:1301.3781}.

\bibitem[\protect\citename{Mikolov \bgroup et al.\egroup
  }2013b]{mikolov2013distributed}
Tomas Mikolov, Ilya Sutskever, Kai Chen, Greg~S Corrado, and Jeff Dean.
\newblock 2013b.
\newblock Distributed representations of words and phrases and their
  compositionality.
\newblock In {\em Advances in Neural Information Processing Systems}, pages
  3111--3119.

\bibitem[\protect\citename{Mikolov \bgroup et al.\egroup
  }2013c]{mikolov2013linguistic}
Tomas Mikolov, Wen-tau Yih, and Geoffrey Zweig.
\newblock 2013c.
\newblock Linguistic regularities in continuous space word representations.
\newblock In {\em Proceedings of NAACL-HLT}, pages 746--751.

\bibitem[\protect\citename{Milajevs and Purver}2014]{milajevs-purver:2014:CVSC}
Dmitrijs Milajevs and Matthew Purver.
\newblock 2014.
\newblock Investigating the contribution of distributional semantic information
  for dialogue act classification.
\newblock In {\em Proceedings of the 2nd Workshop on Continuous Vector Space
  Models and their Compositionality (CVSC)}, pages 40--47, Gothenburg, Sweden,
  April. Association for Computational Linguistics.

\bibitem[\protect\citename{Mitchell and Lapata}2008]{mitchell2008vector}
Jeff Mitchell and Mirella Lapata.
\newblock 2008.
\newblock Vector-based models of semantic composition.
\newblock In {\em Proceedings of ACL-08: HLT}, pages 236--244. Association for
  Computational Linguistics.

\bibitem[\protect\citename{Mitchell and Lapata}2010]{lapata2010}
Jeff Mitchell and Mirella Lapata.
\newblock 2010.
\newblock Composition in distributional models of semantics.
\newblock {\em Cognitive Science}, 34(8):1388--1439.

\bibitem[\protect\citename{Montague}1970]{montague1970universal}
Richard Montague.
\newblock 1970.
\newblock Universal grammar.
\newblock {\em Theoria}, 36(3):373--398.

\bibitem[\protect\citename{Polajnar and Clark}2014]{polajnar-clark:2014:EACL}
Tamara Polajnar and Stephen Clark.
\newblock 2014.
\newblock Improving distributional semantic vectors through context selection
  and normalisation.
\newblock In {\em Proceedings of the 14th Conference of the European Chapter of
  the Association for Computational Linguistics}, pages 230--238, Gothenburg,
  Sweden, April. Association for Computational Linguistics.

\bibitem[\protect\citename{{\v R}eh{\r u}{\v r}ek and Sojka}2010]{rehurek_lrec}
Radim {\v R}eh{\r u}{\v r}ek and Petr Sojka.
\newblock 2010.
\newblock {Software Framework for Topic Modelling with Large Corpora}.
\newblock In {\em {Proceedings of the LREC 2010 Workshop on New Challenges for
  NLP Frameworks}}, pages 45--50, Valletta, Malta, May. ELRA.
\newblock \url{http://is.muni.cz/publication/884893/en}.

\bibitem[\protect\citename{Sch{\"u}tze}1997]{schutze1997ambiguity}
Hinrich Sch{\"u}tze.
\newblock 1997.
\newblock Ambiguity resolution in natural language learning. csli.
\newblock {\em Stanford, CA}, 4:12--36.

\bibitem[\protect\citename{Socher \bgroup et al.\egroup
  }2012]{socher2012semantic}
Richard Socher, Brody Huval, Christopher~D Manning, and Andrew~Y Ng.
\newblock 2012.
\newblock Semantic compositionality through recursive matrix-vector spaces.
\newblock In {\em Proceedings of the 2012 Joint Conference on Empirical Methods
  in Natural Language Processing and Computational Natural Language Learning},
  pages 1201--1211. Association for Computational Linguistics.

\bibitem[\protect\citename{Stolcke \bgroup et al.\egroup }2000]{Stolcke.etal00}
Andreas Stolcke, Klaus Ries, Noah Coccaro, Elizabeth Shriberg, Rebecca Bates,
  Daniel Jurafsky, Paul Taylor, Carol~Van Ess-Dykema, Rachel Martin, and Marie
  Meteer.
\newblock 2000.
\newblock Dialogue act modeling for automatic tagging and recognition of
  conversational speech.
\newblock {\em Computational Linguistics}, 26(3):339--373.

\bibitem[\protect\citename{Turney \bgroup et al.\egroup
  }2010]{turney2010frequency}
Peter~D Turney, Patrick Pantel, et~al.
\newblock 2010.
\newblock From frequency to meaning: Vector space models of semantics.
\newblock {\em Journal of artificial intelligence research}, 37(1):141--188.

\bibitem[\protect\citename{Turney}2006]{turney2006similarity}
Peter~D Turney.
\newblock 2006.
\newblock Similarity of semantic relations.
\newblock {\em Computational Linguistics}, 32(3):379--416.

\bibitem[\protect\citename{Webb \bgroup et al.\egroup }2005]{webb2005dialogue}
Nick Webb, Mark Hepple, and Yorick Wilks.
\newblock 2005.
\newblock Dialogue act classification based on intra-utterance features.
\newblock In {\em Proceedings of the AAAI Workshop on Spoken Language
  Understanding}. Citeseer.

\end{thebibliography}
